\documentclass[10pt,twocolumn,letterpaper]{article}
\PassOptionsToPackage{table}{xcolor}
\usepackage{cvpr}              

\newcommand{\encoder}{\mathbf{E}}

\usepackage{graphicx}
\usepackage{amsmath}
\usepackage{amssymb}
\usepackage{booktabs}
\usepackage{multirow}
\usepackage{placeins}
\usepackage{makecell}
\usepackage{bbding} 
\usepackage{svg}
\usepackage{enumitem}
\usepackage{bm}      

\usepackage{blindtext}
\usepackage{bbm}
\usepackage{letltxmacro,xparse}
\LetLtxMacro{\blindtextblindtext}{\blindtext}
\LetLtxMacro{\blindtextBlindtext}{\Blindtext}
\RenewDocumentCommand{\blindtext}{O{\value{blindtext}}}{%
  \begingroup\color{gray}\blindtextblindtext[#1]\endgroup
}
\RenewDocumentCommand{\Blindtext}{O{\value{blindtext}}O{\value{Blindtext}}}{%
  \begingroup\color{gray}\blindtextBlindtext[#1][#2]\endgroup
}

\renewcommand{\numberline}[1]{%
  \ifnum#1<10 
    \makebox[1.3em][l]{#1} 
  \else
    \makebox[1.8em][l]{#1} 
  \fi
}

\definecolor{cvprblue}{rgb}{0.21,0.49,0.74}
\usepackage[pagebackref,breaklinks,colorlinks,allcolors=cvprblue]{hyperref}

\def\paperID{18} 
\def\confName{EarthVision}
\def\confYear{2025}

\title{SARFormer -- An Acquisition Parameter Aware Vision Transformer for Synthetic Aperture Radar Data}

\author{Jonathan Prexl$^*$ \qquad Michael Recla$^*$ \qquad Michael Schmitt \\
Department of Aerospace Engineering \\
University of the Bundeswehr Munich, Germany \\
{\tt\small \{jonathan.prexl, michael.recla, michael.schmitt\}@unibw.de} \\
{\small $^*$These authors contributed equally to this work }
}

\begin{document}
\maketitle

\begin{abstract}
This manuscript introduces \textit{SARFormer}, a modified Vision Transformer (ViT) architecture designed for processing one or multiple synthetic aperture radar (SAR) images. Given the complex image geometry of SAR data, we propose an acquisition parameter encoding module that significantly guides the learning process, especially in the case of multiple images, leading to improved performance on downstream tasks. We further explore self-supervised pre-training, conduct experiments with limited labeled data, and benchmark our contribution and adaptations thoroughly in ablation experiments against a baseline, where the model is tested on tasks such as height reconstruction and segmentation. Our approach achieves up to $17 \%$ improvement in terms of RMSE over baseline models.
\end{abstract}

\addtocontents{toc}{\protect\setcounter{tocdepth}{-1}}
\section{Introduction}
\label{sec:introduction}

\begin{figure}
    \centering
    \includegraphics[width=\linewidth]{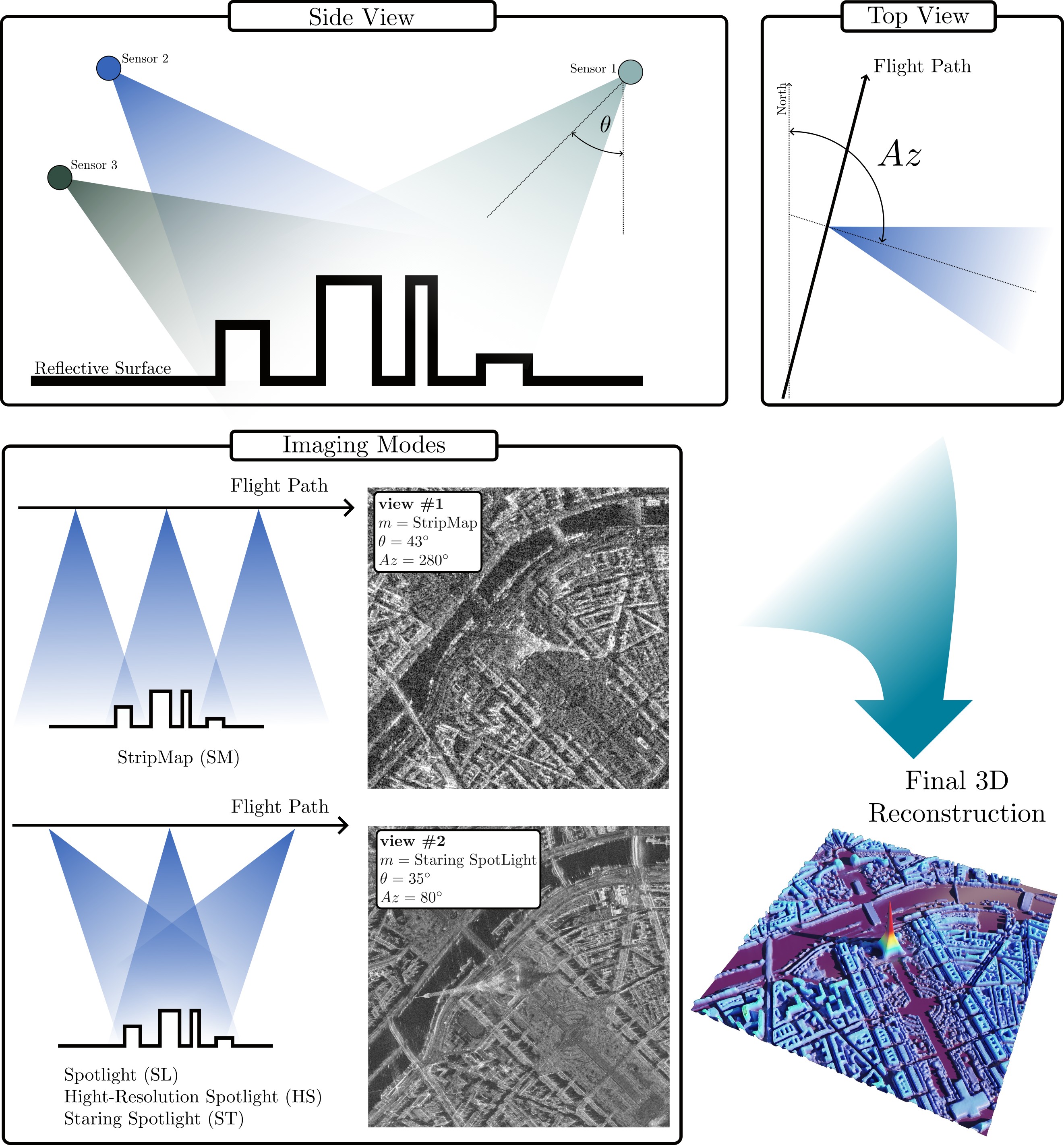}
    \caption{Schematic illustration of the SAR imaging acquisition geometry and modes. Top left: Plane along the flight direction, showing the viewing angle $\theta$ for both descending and ascending orbits. Top right: Azimuth angle $Az$, measured relative to true north. Bottom left: Acquisition modes for a SAR sensor, each accompanied by a visual example demonstrating data captured over the Eiffel Tower. Bottom right: Example output of one of the downstream tasks (height reconstruction) using the proposed \mbox{\textit{SARFormer}} architecture.}
    \label{fig:sar_intro}
\end{figure}

Data acquired by satellite-based Earth observation (EO) missions plays a crucial role in addressing many scientific and socially impactful issues \cite{tassa2020impact}.
In addition to multi- and hyperspectral cameras, another important class of sensors is represented by the active sensing technique known as \textit{Synthetic Aperture Radar} (SAR). 
Sensors utilizing this sensing principle exhibit fundamental differences. SAR operates by measuring the reflections of actively emitted microwave pulses. Since microwaves experience minimal atmospheric interference, this enables a sensing technique that is independent of weather and daytime. This capability is particularly valuable for time-critical applications, where optical sensors may be hindered by cloud cover or insufficient lighting. Additionally, the long wavelengths enable the retrieval of information such as surface roughness and soil moisture.
Moreover, SAR inherently offers insights into the Earth's three-dimensional structure through its range measurement principle.

Even though SAR data is a valuable data source for many applications, on the hind side, SAR images are commonly seen as complex to interpret, mostly due to their imaging geometry and the geometrical effects that come with it, such as layover, foreshortening, and radar shadows \cite{sarhandbook}.
These effects lead to non-intuitive structures, caused by mixed signals from equidistant backscatterers and occlusions, which appear misleading to an untrained observer. 
To correctly interpret a SAR image and the occurring distortions, an understanding of the imaging geometry is of the highest importance.
Within this work, we will refer to three major acquisition variables, the \textit{looking angle} $\theta$, \textit{azimuth angle} $Az$ and \textit{imaging mode} $m$ as the acquisition parameters.
\Cref{fig:sar_intro} gives a graphical illustration of the geometrical meaning of those acquisition parameters. 
While monitoring a certain location on the Earth's surface over time under different viewing geometries, one obtains $N_v$ SAR views $\mathbf{x}_v$ with $v \in (1,\hdots,N_v)$ together with the corresponding acquisition parameters $(\theta_v, Az_v, m_v )$ for each view $v$.

Given all the above-stated information, one can derive the importance of adaptations to deep-learning-based information processing to derive geospatial products from SAR data in a sensor-aware manner.
In this manuscript, we propose a modified ViT-based architecture and pre-training strategy which we refer to as \textit{SARFormer} for the processing of SAR backscatter data, under the incorporation of the relevant acquisition parameters. 
The essence of our contribution can be summarized as follows:
\begin{itemize}[itemsep=0.75pt,topsep=3pt]
    \item We pre-train large transformer-based models on SAR data samples from various geometric conditions using a masked autoencoder, testing for domain-specific pre-training masking strategies to account for the unique data properties.
    \item To pre-train and subsequently fine-tune the models on data with varying acquisition geometries, we introduce an \textit{Acquisition Parameter Encoding} module, which ensures the incorporation of all relevant information for correctly interpreting the scene. 
    \item We test the effect of our proposed contributions in a multi-task setting on three downstream tasks that require detailed image understanding in order to interpret the corresponding SAR scenes.
\end{itemize}

\begin{figure*}
    \centering
    \includegraphics[width=0.91\linewidth]{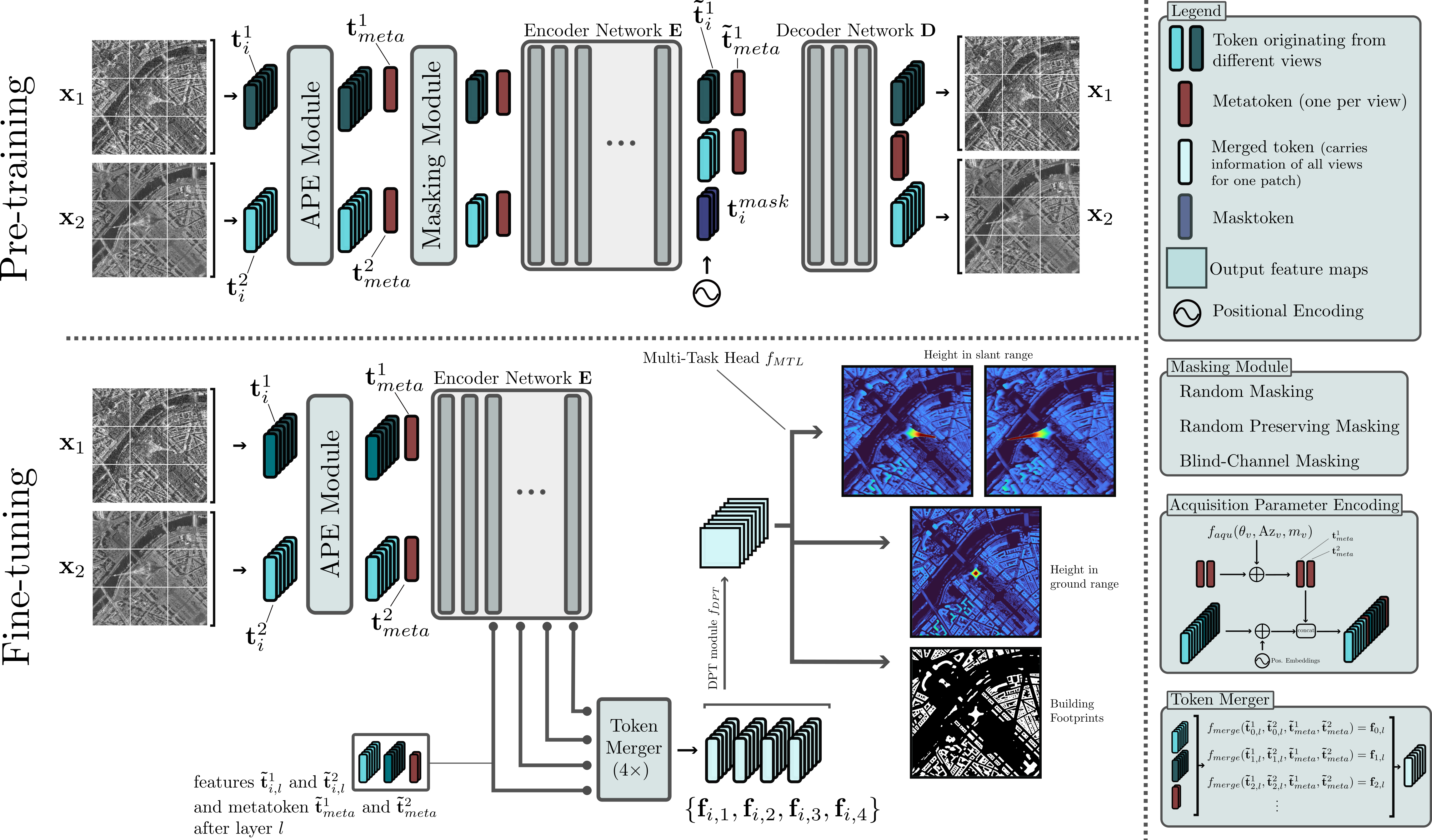}
    \caption{The proposed architecture modifications and the corresponding nomenclature for the pre-training scenario (top) and the fine-tuning stage carried out in a multi-task manner. The two scenarios drawn here correspond to a two-view case but can be generalized to one or multiple views as done in the experimental section of this work.}
    \label{fig:architecture}
\end{figure*}

\section{Related Work}
\label{sec:relatedwork}

\noindent\textbf{Vision Transformer for image understanding} In recent years, the computer vision community has seen a shift towards the use of Transformer architectures, which were initially pioneered for natural language processing tasks. Vision Transformers (ViTs) \cite{dosovitskiy2021imageworth16x16words} have demonstrated remarkable performance on a variety of image understanding problems by effectively capturing long-range dependencies through self-attention mechanisms. Building upon the success of ViTs, the Dense Prediction Transformer (DPT) \cite{Ranftl2021dpt} further extends this paradigm to dense prediction tasks, such as semantic segmentation and depth estimation, making it widely used \cite{oquab2024dinov2,you2023diffusion,yang2024depthanything,bhat2023zoedepth}. 

\vspace{1mm}
\noindent\textbf{Pre-training and Foundation Models in Remote Sensing} ViTs have also gained traction in remote sensing, not least due to their effective pre-training capabilities. Decades of Earth observation have generated abundant data, but high-quality labels remain scarce for many downstream tasks. Self-supervised learning on these vast datasets enables the development of powerful so-called foundation models, which aim to be fine-tuned faster and more accurately while requiring fewer labeled samples.
Masked Image Modeling (MIM) is a highly effective pre-training strategy, masking parts of the input image and training the model to predict the missing content. This approach encourages learning meaningful representations of the image's structure and semantics without labeled data. A popular implementation of MIM is the Masked Autoencoder (MAE) \cite{he2022masked}, which we also utilize in this work.
While MIM-based models have shown significant success with optical and multispectral data \cite{sun2023ringmo,wang2023vitmae,jakubik2023fmibm,cong2022satmae,hong2024spectralgpt} and in multi-modal settings \cite{fuller2022satvit,fuller2023croma,guo2024skysense,wang2022dinomm}, their application to very high resolution (VHR) SAR data remains limited. Notable exceptions include works focused on automatic target recognition (ATR) \cite{li2024maesaratr,li2024saratrxbuildingfoundationmodel} and the \textit{MERLIN} framework \cite{dalsasso2024merlinseg}, which uses image despeckling for pre-training in a segmentation task. However, the development of SAR foundation models still lags behind, particularly for VHR data and tasks beyond object detection and recognition.

Recent works have extended flexibility regarding different sensors and settings by introducing a sensor parameter-aware approach for multi-satellite setups \cite{prexl2024senpamaesensorparameteraware}. This allows incorporation of diverse data sources within a single network architecture while preserving physical signal integrity.

\vspace{1mm}
\noindent\textbf{Deep learning-based mapping from SAR} In spite of their exotic nature, deep learning techniques have also gained momentum in the analysis of SAR data in recent years. In this work, we focus on two relevant downstream tasks: height estimation from single images (analogous to single-image depth estimation from side-view optical images) and building footprint mapping. Both tasks are relevant for global urban mapping applications such as environmental monitoring, disaster response management, and reconnaissance. Compared to optical remote sensing, research on these tasks using SAR data remains limited. For example, \cite{recla2024_sar2heightframework} estimates height from single images in image geometry, then transforms to map geometry in post-processing. Alternatively, \cite{recla2024_sar2heightdual} directly predicts height in map geometry from two images, bypassing the intermediate step. This method requires the model to better understand both scene and acquisition geometry. Building segmentation in SAR imagery has also gained attention, notably in the \textit{SpaceNet6} contest \cite{Shermeyer2020SpaceNet6M}, which focused on a limited local airborne dataset. Additionally, \textit{MERLIN} \cite{dalsasso2024merlinseg} segments buildings in image geometry, while \cite{recla2024_sar2footprints} and \cite{memar2024footprintscosmoskymed} segment directly in map geometry. All these methods rely on conventional CNN architectures; some incorporate auxiliary inputs like the looking angle \cite{recla2024_sar2heightframework, recla2024_sar2heightdual}, crucial for height estimation as it directly influences layover extent.

Given the fundamental role of acquisition geometry, especially when transitioning to map geometry, we introduce \textit{SARFormer} -- a modern architecture designed to seamlessly integrate auxiliary parameters. This framework enhances robustness and versatility in interpreting SAR imagery, demonstrating strong generalization across diverse datasets and efficient pre-training capabilities.

\section{Method}
\label{sec:methods}

The objective of the proposed \textit{SARFormer} architecture is to process one or more SAR acquisitions (views) of the same geographical area on the Earth's surface, captured under different observing angles and acquisition modes while maintaining the physical integrity of the signals.
We restrict our work to the processing of focused SAR backscatter signals without the incorporation of the recorded phase information. The goal is to generate widely usable and generalizable representations that can be used for downstream tasks such as height reconstruction, object detection, or semantic segmentation. We therefore test the descriptiveness of the representations in a multitask framework, ensuring a meaningful and broad impact of the proposed adaptations. 

\vspace{1mm}
\noindent\textbf{Nomenclature}
The input data is given by a set of SAR acquisitions (backscatter) denoted by $\mathbf{x}_v \in \mathbb{R}^{1 \times W \times H}$ where $v \in 1, \hdots, N_v$ indicates the view index and $W$ and $H$ the width and height of the image, respectively. Without a loss of generalization, we will restrict $N_v$ to $N_v \in (1,2,3)$ within this work, as a compromise between insightful experiments and dataset restrictions. Each $\mathbf{x}_v$ has a separate set of acquisition parameters (compare \cref{fig:sar_intro}) which we denote as $\Phi_v = (\theta_v, Az_v, m_v)$. Here the acquisition mode $m$ encodes the index $(0,1,2,3)$ of the four acquisition modes \textit{SM}, \textit{SL}, \textit{HS} and \textit{ST} (compare \cref{fig:sar_intro}, and the supplementary materials for a comparison).
Throughout this work, we build on the classical vision transformer framework \cite{dosovitskiy2021imageworth16x16words} which gets used as an encoder network $\encoder$ in order to process the set of $ \{ \mathbf{x}_v \}$, together with the acquisition parameters $ \{ \Phi_v \}$. We generate an $N_p \times N_p$ patch pattern per view, which, after linear embedding into the latent dimension $d$, leads to $\mathbf{t}_i^v$ tokens, where $i$ indicates the position within the image and $v$ the corresponding originating view. As in \cite{dosovitskiy2021imageworth16x16words}, we encode the position of each token in the image through the addition of learnable positional encoding vectors. 
In light of the above-described initial setup, we describe in the following architectural modifications, contributions, and training strategies. 
To follow along, \cref{fig:architecture} gives a graphical overview of the proposed architecture and nomenclature.

\vspace{1mm}
\noindent\textbf{Acquisition Parameter Encoding (APE)}
The specific viewing geometry of a SAR acquisition is highly relevant, especially for the interpretation of the SAR typical image effects such as layover, shadowing, and foreshortening. To incorporate the acquisition parameters $\Phi_v$ into $\encoder$ we utilise a linear layer $f_{aqu}$ which transfers preprocessed acquisition parameters $\Phi_v^{pre}$ into the latent dimension $d$ of the model:
\begin{equation}
    f_{aqu}: \mathbb{R}^4 \xrightarrow{} \mathbb{R}^d, \text{ } \Phi_v^{embed} = f_{aqu}(\Phi_v^{pre})
\end{equation}
where processing is given by (compare supplementary materials for details):
\begin{equation}
    \Phi_v^{pre} = ( \cos(Az),  \sin(Az), 1 / \tan(\theta), m).
    \label{eq:preprocessing}
\end{equation}
For $N_v$ input views, we generate $N_v$ learnable \textbf{\textit{metatokens}} $\mathbf{t}_{\text{meta}}^v \in \mathbb{R}^d$. For each view $v$, we add the corresponding embedded view parameter, resulting in $\mathbf{t}_{\text{meta}}^v + \Phi_v^{\text{embed}}$.
The input to the transformer backbone for a two-view input is hence given by $2 \times N_p\times N_p$ image tokens and $2$ \textit{metatokens}.
We refer to this step as the acquisition parameter encoding (APE).
We further experimented by adding the embedded acquisition parameters,  $\Phi_v^{embed}$, directly to the image tokens or expanding them through concatenation. Among these approaches, the \textit{metatoken} concept proved to be the most efficient, and therefore, it is the only one discussed in the remainder of the manuscript.

\vspace{1mm}
\noindent\textbf{Task Specific Architecture} 
In order to predict pixel-wise quantities, we work within the DPT \cite{Ranftl2021dpt} framework which collects and merges information from multiple layers of $\encoder$. From each layer, we obtain $N_v \times N_p \times N_p + N_v$ tokens as described above. We will denote the encoded tokens after layer $l$ as $\Tilde{\mathbf{t}}_{i,l}^v$ and the \textit{metatoken} after layer $l$ as $\Tilde{\mathbf{t}}_{meta}^v$.
We merge all tokens originating from the same spatial position $i$ as well as the \textit{metatokens} $\mathbf{t}_{meta}^v$ with a linear fully connected layer followed by a GELU activation $f_{merge}$ to obtain one feature $\mathbf{f}_{i,l}$ corresponding to the patch position $i$ for a given layer $l$:
\begin{equation}
\begin{split}
    &f_{merge}: \mathbb{R}^{2 \cdot N_v \cdot d} \xrightarrow{} \mathbb{R}^d, \\
    &\mathbf{f}_{i,l} = f_{merge}(
    \Tilde{\mathbf{t}}_{i,l}^1,
    \Tilde{\mathbf{t}}_{i,l}^2,\hdots,
    \Tilde{\mathbf{t}}_{meta}^1,
    \Tilde{\mathbf{t}}_{meta}^2,\hdots)
\end{split}
\end{equation}
which carries the information merged from all available views.
Hence we obtain $N_p \times N_p$ features of dimension $\mathbb{R}^d$ for each extraction layer $l$.
From here we can directly follow \cite{Ranftl2021dpt} and use features gathered from different layers and processed on different resolution levels which results in $N_{conv}$ feature maps $\mathbf{f}_i^{conv}$ (compare supplementary materials for details).
In order to test the model's capabilities of the 3D understanding, we use $\mathbf{f}_i^{conv}$ to obtain three pixel-wise masks in a multi-task setting. A detailed explanation can be found in \cref{sec:exp}.

\vspace{1mm}
\noindent\textbf{Pre-training} Since the field of EO delivers a huge amount of unlabeled data due to the constant monitoring of the Earth's surface, self-supervised pertaining yields large potential.
We generally follow the \textit{masked autoencoder} (MAE) \cite{he2022masked} strategy to pre-train \textit{SARFormer} in a self-supervised manner.
Since SAR (more general EO data) usually covers large spatial extents due to the moderate image resolution, images are generally not object-centered which can require an adapted masking ratio or strategy. 
While keeping the aforementioned architectural adaptations of \textit{SARFormer} including the APE module, we test for three different masking strategies in the MAE setup. For pre-training with a single SAR acquisition, we apply random masking with no further modification. We implement the same procedure for a two-view case and refer to it as \textit{random}. Further, we conduct a \textit{preserving} masking strategy where for a given patch-position we leave one of the two tokens $\mathbf{t}_i^1$ or $\mathbf{t}_i^2$ unmasked. Filling in the remaining gaps requires an understanding of the relationship of different view geometries $\Phi_v$ but reduces large unmasked areas that cover complete objects and hence can't be reconstructed.
Additionally, we test an extreme version of this by masking all tokens from either view one or two. We refer to this strategy as \textit{blind-channel} masking. 
In order to remain with constant masking values to make experiments comparable, we further apply random masking of the remaining patches for the two strategies \textit{preserving} and \textit{blind-channel} to obtain a fixed masking ratio of $75\%$.

\section{Experimental Setup}
\label{sec:exp}

\noindent\textbf{SAR data} Our dataset comprises over $200$ \mbox{TerraSAR-X} (TSX) satellite acquisitions spanning multiple imaging modes (SM, SL, HS, ST), and different acquisition settings (orbit directions, looking angles). The data covers more than 25 distinct geographical regions, predominantly but not exclusively focusing on urban environments. While the dataset covers different areas from across the globe, it exhibits a geographical bias toward European and North American cities, due to the availability of high-resolution labels for downstream tasks. Each SAR image, provided in Single Look Complex (SLC) format, is geocoded by projecting it onto a globally available digital terrain model (DTM), the \mbox{FABDEM} \cite{hawker2022fabdem}, using the known sensor model and orbits. With a common grid spacing of 1~m on the ground, this geocoding process aligns the images, enabling them to be directly compared across different acquisition settings and spatial resolutions. The terrain model, intentionally devoid of man-made structures and vegetation, ensures that ground points captured from varying angles map consistently onto the same locations. However, elevated objects -- such as buildings -- appear displaced, which is typically considered a parallax error. Here, this displacement effect, as illustrated by the Eiffel Tower example in \cref{fig:sar_intro}, is leveraged as a measurement signal. The complex pixel values are calibrated to backscatter intensities, then clipped to a range of -30 to +10~dB, and subsequently normalized to the $[0,1]$ interval for neural network input.

\vspace{1mm}
\noindent\textbf{Labels} The dataset’s labels incorporate three types of ground truth data for multi-task learning: building footprints, heights in a map projection system (e.g., UTM), and heights in the image-specific slant geometry. 
Building footprint labels result from fusing OpenStreetMap (OSM) with Microsoft’s Global ML Building Footprints, favoring OSM for its manual curation in overlapping areas. This fused dataset was manually inspected and masked to remove apparent errors.
Height annotations are derived from publicly available LiDAR data provided by regional surveying authorities. Since height labels are not available for all cities in the SAR imagery, some locations lack complete ground truth information. Absolute elevations are converted to heights above ground by subtracting a terrain model. The heights are further projected into the image geometries (\textit{slant}) following \cite{recla2022} and then, through the geocoding procedure, aligned with the SAR images.
In layover regions -- where multiple targets share a resolution cell -- the highest point's height is mapped, while SAR shadow occlusions remain unaddressed. Refer to the three example outputs depicted in Fig.~\ref{fig:architecture}, or to the supplementary materials, for a visualization.

\vspace{1mm}
\noindent\textbf{Sampling Strategy} The matching pairs of SAR images, height data (both in map and image projection), and building footprints are cropped into tiles with a size of 384×384 pixels. For each tile, the relevant metadata -- including the looking angle and azimuth angle at the center pixel, as well as the imaging mode -- is stored alongside the image data. Due to the substantially larger spatial coverage of SM scenes compared to SpotLight images, the number of SM patches significantly exceeds those from the other modes. We employ weighted random sampling to address this imbalance, ensuring that approximately equal numbers of patches from each imaging mode are presented to the model in each epoch. Additionally, each epoch includes 25\% of samples without corresponding height data, allowing the model to encounter all locations, even those lacking height labels. For these specific patches, the height estimation is excluded from the height loss calculation. 

\begin{table*}[t!]
    \centering
    \caption{Numerical results of the fully-supervised experiments (no pre-training). Best values are evaluated according to the number of views. $\dagger$ indicates values taken from other, not publicly available publications (test datasets and input data differ, so only limited comparable). All metrics of \textit{SARFormer} were generated using the ViT-Large configuration.}
    \resizebox{.7\linewidth}{!}{
    \begin{tabular}{cccccccccc}
    \toprule
    \multicolumn{2}{c}{} & \multicolumn{2}{c}{\textbf{\makecell{Classification \\ Footprints}}} & \multicolumn{3}{c}{\textbf{\makecell{Regression \\ Height}}} & \multicolumn{3}{c}{\textbf{\makecell{Regression \\ Height (Slant)}}}\\
    \cmidrule(lr){3-4} \cmidrule(lr){5-7} \cmidrule(lr){8-10} 
    \# Views & Model  & mIoU & OA & MAE & RMSE & SSIM & MAE & RMSE & SSIM \\
    \cmidrule(lr){1-10} 
    \multirow{5}{*}{1} & \footnotesize{SAR2Footprints} \cite{recla2024_sar2footprints} & \footnotesize{$0.61^\dagger$} & \footnotesize{$0.78^\dagger$} & - & - & - & - & - & - \\
     & \footnotesize{SAR2Height} \cite{recla2024_sar2heightframework} & - & - & \footnotesize{$5.60^\dagger$} & \footnotesize{$8.30^\dagger$} & - & \footnotesize{$3.50^\dagger$} & - & \footnotesize{$0.84^\dagger$} \\
    \arrayrulecolor{black!30}\cmidrule(lr){2-10}
                       & UNet MTL \cite{ronneberger2015} & \bm{$0.68$} & \bm{$0.90$} & $5.67$ & $8.55$ & \bm{$0.82$} & $5.39$ & $7.80$ & \bm{$0.84$} \\
                       & DPT-Large \cite{Ranftl2021dpt} & $0.67$ & $0.89$ & $5.45$ & $7.89$ & \bm{$0.82$} & $5.35$ & $7.70$ & \bm{$0.84$} \\
                       & SARFormer (ours) & $0.67$ & $0.89$ & \bm{$5.24$} & \bm{$7.61$} & \bm{$0.82$} & \bm{$5.29$} & \bm{$7.60$} & \bm{$0.84$} \\
    \arrayrulecolor{black} \cmidrule(lr){1-10}
    \multirow{4}{*}{2} & \footnotesize{SAR2Height-Dual \cite{recla2024_sar2heightdual}} & - & - & \footnotesize{$3.67^\dagger$} & - & \footnotesize{$0.87^\dagger$} & - & - & - \\
    \arrayrulecolor{black!30} \cmidrule(lr){1-10}
                       & UNet MTL \cite{ronneberger2015} & \bm{$0.74$} & \bm{$0.92$} & $4.77$ & $7.35$ & $0.84$ & \bm{$4.64$} & \bm{$6.89$} & \bm{$0.88$} \\
                       & DPT-Large \cite{Ranftl2021dpt} & $0.73$ & \bm{$0.92$} & $4.61$ & $6.87$ & $0.85$ & $4.77$ & $6.99$ & \bm{$0.88$} \\
                       & SARFormer (ours) & \bm{$0.74$} & \bm{$0.92$} & \bm{$4.39$} & \bm{$6.60$} & \bm{$0.85$} & $4.69$ & $6.90$ & \bm{$0.88$} \\
    \arrayrulecolor{black} \cmidrule(lr){1-10}
    \multirow{3}{*}{3} & UNet MTL \cite{ronneberger2015} & \bm{$0.74$} & \bm{$0.92$} & $4.95$ & $7.57$ & $0.85$ & $4.85$ & $7.11$ & $0.88$ \\
                       & DPT-Large \cite{Ranftl2021dpt} & \bm{$0.74$} & \bm{$0.92$} & $4.63$ & $6.90$ & $0.85$ & $4.88$ & $7.15$ & $0.88$ \\
                       & SARFormer (ours) & \bm{$0.74$} & \bm{$0.92$} & \bm{$4.30$} & \bm{$6.39$} & \bm{$0.86$} & \bm{$4.43$} & \bm{$6.52$} & \bm{$0.89$} \\
    \arrayrulecolor{black}\bottomrule
    \end{tabular}
    }
    \label{tab:fromScratch}
\end{table*}


\vspace{1mm}
\noindent\textbf{Fully supervised training} To evaluate the proposed APE module and compare the \textit{SARFormer} architecture to our baseline setting, we trained models with one, two, and three input views, comparing configurations with and without the APE module (\textit{metatoken}). The primary experimental setup employed the ViT-Large architecture \cite{dosovitskiy2021imageworth16x16words}, characterized by $24$ transformer layers and $16$ attention heads. While supplementary experiments explored ViT-Base and ViT-Huge, ViT-Large was selected as the primary model due to its preferable balance between computational efficiency and performance. Comparisons to the other model sizes can be found in the supplementary material.
The training protocol consisted of 250 epochs using the AdamW optimizer, with a half-cycle cosine decay learning rate scheduler preceded by a three-epoch warm-up period. Each training epoch incorporated 100,000 randomly chosen balanced samples (of which 25,000 were without height labels). The loss function for the regression task (height estimation)  was designed as a composite metric, combining asymmetric L1 loss, gradient loss, and normal loss \cite{hu2018loss}. For the segmentation task, binary cross-entropy loss was used. The total loss function is formulated as a weighted sum of the individual task-specific losses. Details can be found in the supplementary materials.

\vspace{1mm}
\noindent\textbf{Pre-Training} For both the single-view and two-view cases, backbones were pre-trained using masked autoencoders. Their decoders were kept intentionally shallow, with only three layers, to ensure that most of the descriptive features are learned in the encoders. In the two-view scenario, the three previously described masking strategies were applied to evaluate their effects on the learning capacity of the network across multiple perspectives. Each MAE model was trained over 1500 epochs with 100,000 balanced samples per epoch using the masked $\mathcal{L}_1$ loss function on the reconstructed images. It is important to note that the single-view case benefits from a larger set of unique training samples, as it is not constrained by the requirement of having a second view of the same area.

\vspace{1mm}
\noindent\textbf{Fine-Tuning} With the pre-trained encoders serving as backbones, \textit{SARFormer} is fine-tuned for both single-view and two-view cases. In the frozen-weight setting, the first two-thirds of the backbone layers are not further trained, which, in the ViT-Large configuration, leaves eight layers unfrozen. Additionally, a fine-tuning experiment without frozen layers is conducted. For the single-view case, the only available masking strategy (random) is applied, while for the two-view case, the impact of the different masking strategies is evaluated. All other settings remain consistent with those used in training the models from scratch.

\vspace{1mm}
\noindent\textbf{Evaluation Dataset} To compare the performance of the different models, we selected two cities with distinct architectural characteristics -- Berlin, Germany, and Vancouver, Canada -- as test regions. These cities were entirely excluded from the training and validation set. Achievable accuracy is highly dependent on the imaging modes used, which determine the spatial resolution of the input images, as well as on the characteristics of the scene itself, such as building density, building heights (and consequently the number of mixed signals in layover areas), vegetation, facade materials, and other factors. These elements affect the signal responses and thus the complexity of accurately interpreting them. Additionally, we expect that multiple views (the more diverse, the richer the information) will enhance the model’s 3D reconstruction capabilities. To ensure a fair comparison, we prepared six test image configurations per city, each tailored to the same region and including all available imaging modes with a wide range of acquisition settings (angles). For each model, we generated a total of 12 output sets (including building footprints, heights in UTM coordinates, and projected slant heights) over which we computed the relevant error metrics for each task. These metrics include overall accuracy (OA) and mean intersection over union (mIoU) for building footprints, as well as root mean squared error (RMSE), mean absolute error (MAE), and structural similarity index measure (SSIM) \cite{Wang2004ssim} for the two height estimation tasks. For each model, the best-performing checkpoint on the validation set was used to generate the error metrics.

\vspace{1mm}
\noindent\textbf{Baselines} Deep learning-based scene interpretation from high-resolution SAR data is an emerging field with no publicly available models or benchmark datasets, which complicates fair comparisons. This is primarily driven by the licensing of data for missions such as TSX, which academic institutions can access but not publicly release.
Nevertheless, this paper aims to put the results in the context of existing work and provide the possibility of benchmarking future approaches as effectively as possible. Therefore, two steps are taken.  
First, we provide all scene IDs used for pre-training as well as detailed evaluation settings (as described above), allowing any research institution with access to the TSX archive to reproduce the results. Secondly, existing works \cite{recla2024_sar2heightframework,recla2024_sar2footprints,recla2024_sar2heightdual} that report metrics on any of the tasks in our multi-task setup are shown in the corresponding tables, even though -- due to differences in the evaluation area -- they are not directly comparable but still can serve as a relative anchor point.  
For a direct comparison of performance, we train two benchmark models: a UNet with a multi-task extension and a vanilla ViT with the DPT \cite{Ranftl2021dpt} extension. This ensures both the validity of our proposed adaptation to the ViT-based architecture as well as the performance enhancement compared to the CNN-based architectures that have been used previously.

\begin{table*}[t!]
    \centering
    \caption{Achieved metrics through pre-training and subsequent fine-tuning. Best values are evaluated according to the used pre-training method (masking strategy). The single-view case allows only for the random masking strategy. For the frozen {\footnotesize \Snowflake} setting, $2/3$ of the backbone's layers were not updated during fine-tuning.}
    \resizebox{.75\linewidth}{!}{
    \begin{tabular}{cclcccccccc}
    \toprule
    \multicolumn{3}{c}{} & \multicolumn{2}{c}{\textbf{\makecell{Classification \\ Footprints}}} & \multicolumn{3}{c}{\textbf{\makecell{Regression \\ Height}}} & \multicolumn{3}{c}{\textbf{\makecell{Regression \\ Height (Slant)}}}\\
    \cmidrule(lr){4-5} \cmidrule(lr){6-8} \cmidrule(lr){9-11} 
    \# Views & \makecell{Frozen \\ Weights} & \makecell{Pre-training \\ Method} & mIoU & OA & MAE & RMSE & SSIM & MAE & RMSE & SSIM \\
    \cmidrule(lr){1-11}
    \multirow{2}{*}{1}  & \footnotesize{\Snowflake} & Random & $0.64$ & $0.88$ & $5.84$ & $8.37$ & $0.81$ & $5.96$ & $8.46$ & $0.83$ \\
    \arrayrulecolor{black!30}\cmidrule(lr){2-11}  
     & & Random & $0.69$ & $0.90$ & $5.07$ & $7.40$ & $0.83$ & $4.97$ & $7.20$ & $0.84$ \\
    \arrayrulecolor{black!30}\cmidrule(lr){1-11}
    \multirow{6}{*}{2}  & \multirow{3}{*}{\footnotesize{\Snowflake}} & Random & $0.71$ & \bm{$0.91$} & \bm{$4.49$} & \bm{$6.59$} & \bm{$0.85$} & \bm{$4.36$} & \bm{$6.41$} & \bm{$0.88$} \\
                        &                     & Blind Channel & $0.71$ & \bm{$0.91$} & $4.61$ & $6.89$ & \bm{$0.85$} & $4.74$ & $6.99$ & $0.87$ \\
                        &                     & Preserving & \bm{$0.72$} & \bm{$0.91$} & $4.65$ & $6.85$ & \bm{$0.85$} & $4.75$ & $6.93$ & \bm{$0.88$} \\
    \arrayrulecolor{black!30}\cmidrule(lr){2-11}  
                        & \multirow{3}{*}{} & Random & \bm{$0.75$} & \bm{$0.92$} & $4.19$ & $6.29$ & $0.85$ & $4.07$ & $6.12$ & \bm{$0.89$} \\
                        &                     & Blind Channel & $0.73$ & \bm{$0.92$} & $4.38$ & $6.56$ & \bm{$0.86$} & $4.53$ & $6.72$ & $0.88$ \\
                        &                     & Preserving & $0.74$ & \bm{$0.92$} & \bm{$4.12$} & \bm{$6.13$} & \bm{$0.86$} & \bm{$3.96$} & \bm{$5.90$} & \bm{$0.89$} \\

    \arrayrulecolor{black}\bottomrule
    \end{tabular}
    }
    \label{tab:pretrained}
\end{table*}

\section{Results}
\label{sec:results}

In this section, we empirically investigate the effects of the two earlier introduced adaptations, namely the APE module and the domain-adapted pre-training strategies given by the three different masking functions. As earlier mentioned, we test the descriptiveness of the features obtained by the \textit{SARFormer} encoder $\mathbf{f_i} = \encoder(\mathbf{x}_v,\Phi_v)$ by linking up to a multi-task setting where pixel-wise height above ground is estimated in the map and image geometry and building footprints are predicted in a binary segmentation setting. We compare the contribution and effects of each introduced adaptation and provide an absolute comparison to self-generated baseline results using a vanilla CNN and ViT-based deep-learning approach. 

\vspace{1mm}
\noindent\textbf{Effect of APE} To evaluate the impact of the proposed APE module (\textit{metatoken}), we initially conducted experiments without pre-training. In \cref{tab:fromScratch}, we present a comparison of the \textit{SARFormer} model with the APE module enabled, alongside the baseline models as a function of the number of input views $N_v$.
Examining the effect of the APE module, we observe that it consistently improves or maintains performance across all metrics. Additionally, \mbox{\textit{SARFormer}} outperforms baseline models in nearly all configurations; however, this advantage is less evident for classification tasks, where results remain comparable. 

\vspace{1mm}
\noindent\textbf{Effect of Pre-training Strategies} \Cref{tab:pretrained} presents the results for models trained with the three proposed MAE-based pre-training strategies. Experiments were limited to configurations with one or two SAR views, and we tested two scenarios for each: fine-tuning all model weights and fine-tuning only the last $1/3$ of the ViT layers while keeping the remainder of the weights frozen. The results indicate that the proposed \textit{preserving} masking strategy achieves superior performance across most metrics, suggesting its suitability for non-object-centered remote sensing imagery (compare \cref{sec:methods}). In contrast, the \textit{blind-channel} strategy, representing an extreme case of \textit{preserving}, underperformed, likely due to the increased task complexity and the model's limited capacity to extract robust features in this configuration.
All pre-trained and fine-tuned models in \Cref{tab:pretrained} utilized the APE module, enabling comparison with training from scratch in \cref{tab:fromScratch}. In the optimal configuration (\textit{preserving} pre-training with full fine-tuning), we observe a 2-point gain in IoU and a 4-6\% improvement in MAE compared to the from-scratch baselines.

These improvements are even more pronounced when training data is limited, as label scarcity can significantly affect the quality of results. We fine-tuned the model using only two labeled SAR images from Paris, introducing significant domain shifts in multiple regards during testing. Despite reduced performance compared to using the full dataset, the benefits of pre-training were evident, highlighting its value in few-label or out-of-domain scenarios. \cref{tab:limited_labels} shows the error metrics obtained. A visual example can be found in \cref{fig:pretrained_fewlabels} in the supplementary materials.

\begin{table}[t!]
    \centering
    \caption{Metrics achieved on the test set using a very limited labeled dataset for training ($2$ SM images from Paris). Pre-training evidently helps for such extreme domain shifts between training and inference.}
    \resizebox{.75\linewidth}{!}{
    \begin{tabular}{lcccc}
    \toprule
    \textbf{Model} & \makecell{\textbf{pre-} \\ \textbf{trained}} & \textbf{mIoU} & \makecell{\textbf{MAE} \\ (map)} &  \makecell{\textbf{MAE} \\ (slant)} \\
    \cmidrule(lr){1-2} \cmidrule(lr){3-3} \cmidrule(lr){4-4} \cmidrule(lr){5-5}
    UNet MTL &  & $0.51$ & $7.61$ & $7.84$ \\
    SARFormer & & $0.54$ & $7.22$ & $8.51$ \\
    SARFormer & \checkmark & \bm{$0.62$} & \bm{$5.66$} & \bm{$6.15$} \\
    \arrayrulecolor{black}\bottomrule
    \end{tabular}
    }
    \label{tab:limited_labels}
\end{table}

\begin{figure}
    \centering
    \includegraphics[width=\linewidth]{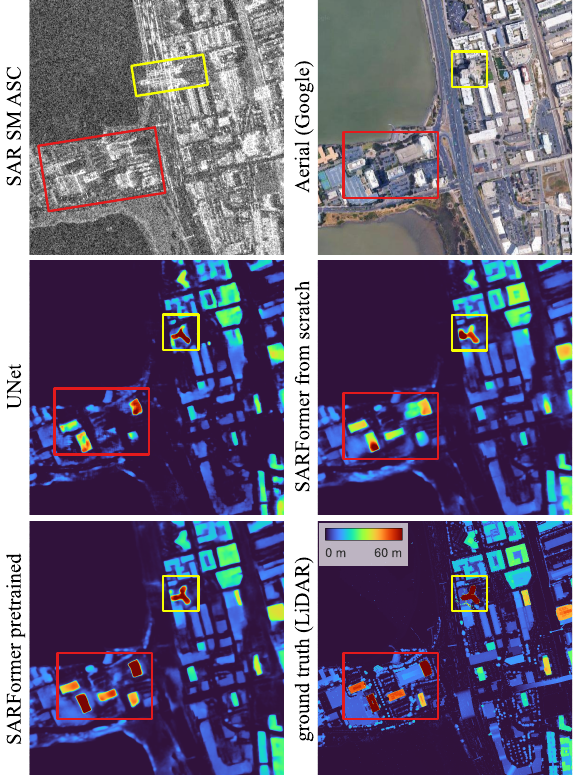}
    \caption{Comparison between the outputs of different models (see labels) next to the input image, an aerial view, and ground truth. Especially building shapes seem to benefit from pre-training. Please find more visual examples in the supplementary materials.}
    \label{fig:highrise_comparison}
\end{figure}

\vspace{1mm}
\noindent\textbf{Qualitative Comparison} Visually, the differences in performance across the models are more pronounced than the numerical error metrics alone suggest. \cref{fig:highrise_comparison} provides a side-by-side comparison of the outputs from four different models, alongside one of the SAR input images and an aerial view for reference. These outputs were generated using a combination of two SM images, a challenging task due to limited spatial resolution. While the height estimates still tend to be underestimated, the \textit{SARFormer} trained from scratch demonstrates a marked improvement in identifying buildings, particularly high-rise structures within the red-marked area. Comparing these results to the pre-trained model output reveals the significant impact of pre-training on both the quality of height estimations and the accuracy of building shapes. Notably, the pre-trained model achieves the most accurate reconstruction of the unusually shaped building highlighted in yellow. More visual examples can be found in the supplementary material.

\vspace{1mm}
\noindent\textbf{Effects of Backbones} 
To assess the impact of model size on performance, we trained our best-performing configuration (2 views, active APE, \textit{preserving} masking during pre-training) with various backbones: ViT-Base, ViT-Large, and ViT-Huge \cite{dosovitskiy2021imageworth16x16words}. While ViT-Huge achieved the highest performance (see supplementary materials), we selected ViT-Large for main experiments against baselines (shown in \cref{tab:fromScratch,tab:pretrained}) due to its computational efficiency and minimal performance trade-off. Results for experiments with the ViT-Base backbone are included in the supplementary materials, demonstrating similar relative performance gains for our adaptations.

\section{Discussion}
\label{sec:discussion}
The performance improvements observed, particularly with pre-training and the integration of the APE module, underscore the effectiveness of domain-specific adaptations for deep learning in remote sensing applications. Notably, the \textit{preserving} masking strategy demonstrated superior capability in capturing useful spatial representations, leading to substantial enhancements in downstream regression and segmentation tasks. Building on the insights from this initial study, future work could explore integrating additional metadata into the APE module -- such as sensor type (optical, SAR, thermal), platform details (source satellite), spatial resolution, frequency band, spectral response functions (like in \cite{prexl2024senpamaesensorparameteraware}), or geographical location -- to increase the model's adaptability across datasets and further advance toward a generalized task-agnostic foundation model for remote sensing data.
\par
Another key strength of \textit{SARFormer}’s Transformer-based architecture is its inherent flexibility with sequence length, allowing it to scale more efficiently and more elegant to multi-view settings compared to traditional convolutional architectures. This adaptability is particularly advantageous in scenarios where multi-acquisition or temporal sequences are required to capture diverse perspectives of a scene. Accordingly, although the performance gain observed in \cref{tab:fromScratch} between the two-view and three-view settings may appear marginal, this is likely due to the limited azimuth diversity in TSX data, which provides either ascending or descending orbit images. Despite the inclusion of a higher-resolution third view (which was intentionally avoided here for testing), it does not contribute much new information to the network. Expanding the dataset to include SAR data from additional orbital paths could provide a more comprehensive, multi-perspective view, very likely enhancing performance in multi-view configurations.
\par
It is important to note that evaluating reconstructed height maps is not as straightforward as it may seem. Geolocation errors, common to most satellite images due to inaccuracies in orbital data or terrain models used in georeferencing, introduce shifts between the labels and predictions, which inflate pixel-wise error metrics. Additionally, discrepancies in the acquisition dates between the label data and SAR imagery lead to differences, particularly in rapidly changing urban areas. Furthermore, with no available benchmark datasets in this emerging field, direct comparison with existing methods is impossible, as achievable accuracy varies significantly with numerous factors, leading to potential biases. Nevertheless, relative comparisons were conducted, and the performance improvements resulting from our proposed adaptations are clearly demonstrated.

\section{Conclusion}
\label{sec:conclusion}

This paper introduces \textit{SARFormer}, a Vision Transformer tailored for VHR SAR data, integrating essential domain knowledge to facilitate the model's scene interpretation. \textit{SARFormer} embeds sensor-specific parameters, such as acquisition angles and modes, directly into its architecture. This allows it to handle geometric distortions in SAR imagery and adapt more effectively across diverse SAR data.
Domain-specific pre-training, particularly using a masked autoencoder (MAE) adapted for remote sensing imagery, proved crucial. The proposed masking strategies enhance feature learning in non-object-centered data, boosting performance in downstream tasks like height estimation and building footprint segmentation.
\textit{SARFormer} offers a step towards a versatile foundation model approach, applicable to multiple tasks. As SAR usage grows in areas like environmental monitoring, disaster response, and surveillance, \textit{SARFormer}’s framework is positioned to support broad and important applications in remote sensing.

{
    \small
    \bibliographystyle{ieeenat_fullname}
    \bibliography{main}
}

\clearpage
\setcounter{page}{1}
\maketitlesupplementary

\tableofcontents
\addtocontents{toc}{\protect\setcounter{tocdepth}{3}} 

\section{Additional Information: Data}

\begin{figure}
    \centering
    \includegraphics[width=\linewidth]{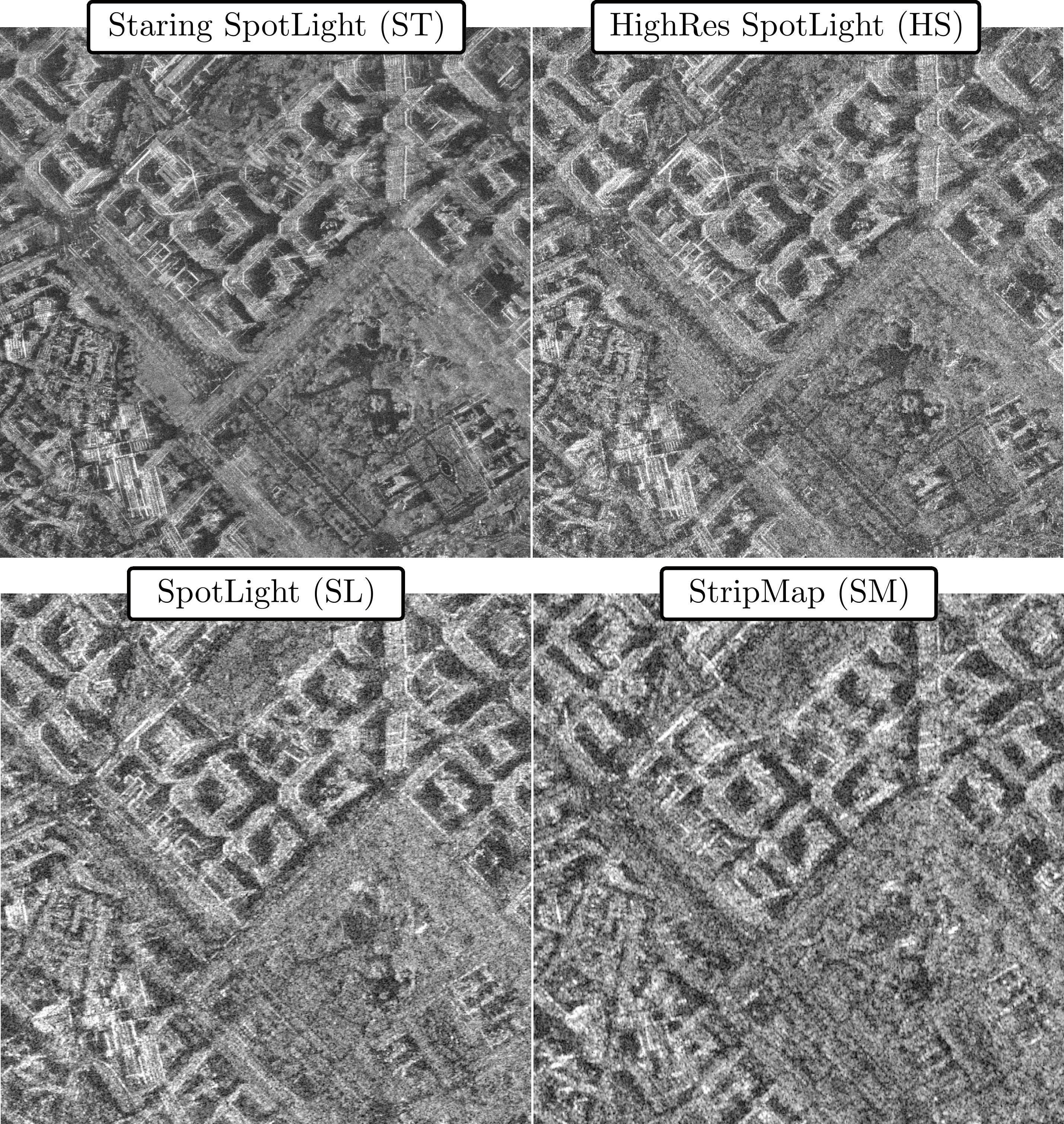}
    \caption{Comparison between the four different imaging modes being used in this work. The images depict the same scene captured with similar viewing angles but in different imaging modes.}
    \label{fig:imaging_modes}
\end{figure}

\noindent\textbf{Imaging Modes}
The TerraSAR-X satellite, like other SAR platforms, offers multiple imaging modes, each providing different spatial resolutions and aerial coverage. In this work, we utilize four imaging modes: \textit{StripMap} (SM), \textit{SpotLight} (SL), \textit{High Resolution SpotLight} (HS), combining both 150 MHz and 300 MHz bandwidths, and \textit{Staring SpotLight} (ST). The key differences among these modes are summarized in \cref{tab:imaging_modes}. It is important to note that a SAR image with lower spatial resolution is not equivalent to a downsampled high-resolution image. This distinction arises primarily because the speckle effect -- a granular noise inherent to coherent imaging systems like SAR, resulting from the constructive and destructive interference of backscattered signals from multiple scatterers within a resolution cell -- has a significantly greater impact on images with larger resolution cells than on fine-resolution SpotLight images. \cref{fig:imaging_modes} compares the same area captured in the four different modes at similar incidence angles (note: different acquisition dates).

\begin{figure*}
    \centering
    \includegraphics[width=\linewidth]{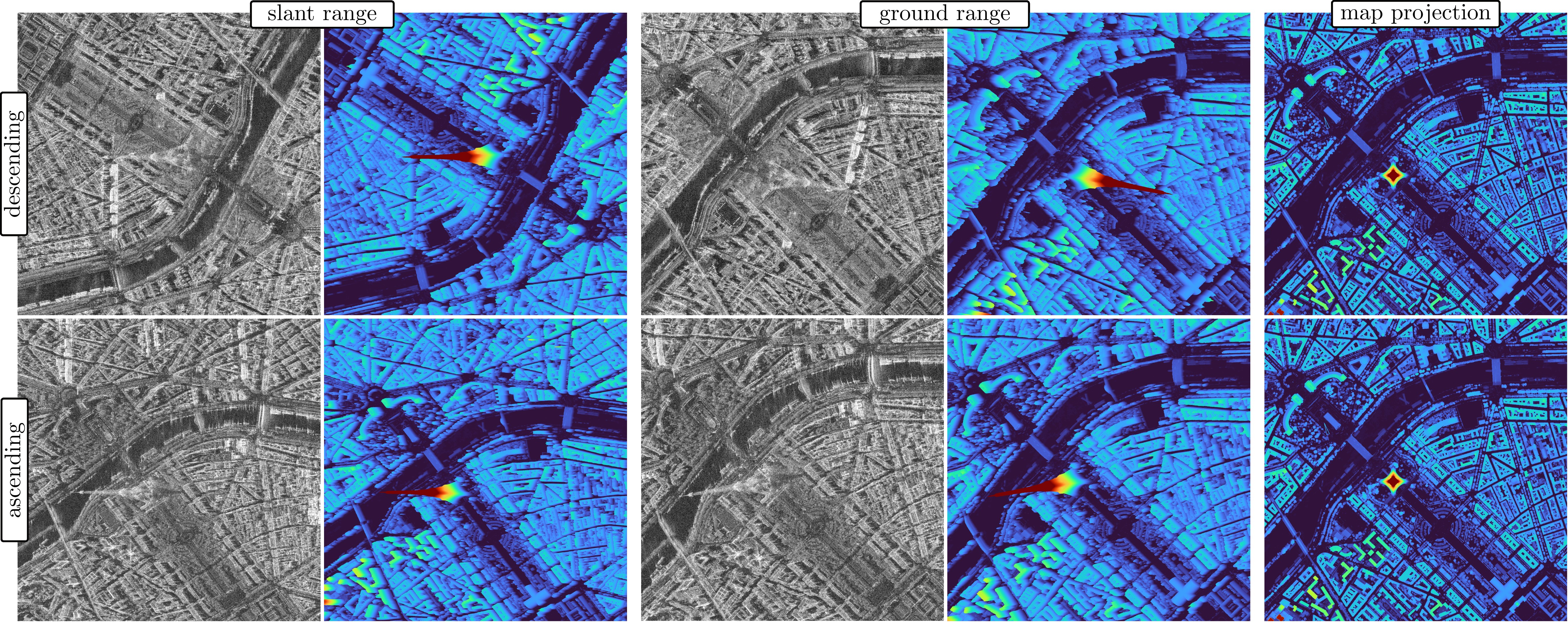}
    \caption{Comparison of the different SAR image geometries. Each row shows a SAR image of the same area taken from different directions, along with the corresponding height above ground values. Columns 1 and 2 display the images in their native slant-range geometry, where the columns represent the distance to the sensor from left to right. In Columns 2 and 3, the images are projected onto a terrain model, making each pixel correspond to one meter on the Earth's surface. The far-right column shows the height values in a map projection, independent of the image geometries, and thus identical for both acquisitions.}
    \label{fig:geometries}
\end{figure*}

\vspace{1mm}
\noindent\textbf{Radar Geometry}
In SAR imaging, two common geometries are used to represent spatial dimensions: \textit{slant range} and \textit{ground range}. The slant range is the direct line-of-sight distance between the radar antenna and a target on the Earth's surface, measured along the radar beam's path. In a typical slant range SAR image, each column represents the slant range distance from the radar to the target, effectively mapping the across-track (range) dimension. Each row corresponds to a different position along the radar platform's flight path, representing the along-track (azimuth) dimension. This means the image grid is organized such that columns increase with distance from the radar, and rows progress with the movement of the radar platform.
Compare the first to columns in \cref{fig:geometries}: One can observe that although the scenes were captured from opposite directions, the top of the Eiffel Tower points to the left side of the image because it is closer to the sensor than the base of the tower. Further, the north direction on the Earth's surface does not correspond to the top in the slant SAR image. Consequently, two SAR images in slant range cannot be accurately coregistered unless they were acquired from the same orbit with identical acquisition parameters.

\begin{table}
	\begin{center}
	\caption{Properties of the different imaging modes of TerraSAR-X as used in this work}
		\label{tab:imaging_modes}
        \resizebox{\linewidth}{!}{
		\begin{tabular}{ccccc}
			\toprule
			\makecell{Imaging \\ Mode} & \makecell{Scene Size \\ \footnotesize{[km]}} & \makecell{Slant Range \\ Resolution \footnotesize{[m]}} & \makecell{Azimuth \\ Resolution \footnotesize{[m]}} &  \makecell{Looking \\ Angle} \\ 
            \cmidrule(r){1-1} \cmidrule(lr){2-2} \cmidrule(lr){3-3} \cmidrule(lr){4-4} \cmidrule(lr){5-5}
			SM &  30 x 50 & 1.2 & 3.3 & 20° -- 45° \\ \addlinespace
			SL &  10 x 10 & 1.2 & 1.7 & 20° -- 55° \\ \addlinespace
			HS &  10 x 5 & 0.6 -- 1.2 & 1.1 & 20° -- 55° \\ \addlinespace
			ST & 4 x 3.7  & 0.6 & 0.24  &  20° -- 45° \\ 
			\bottomrule
		\end{tabular}
        }
	\end{center}
\end{table}

\par
On the other hand, in ground range geometry, these slant range measurements are projected onto a horizontal plane (or, as in this case, a terrain model), as if viewed from directly above. By removing elevation-induced distortions, ground range images simplify spatial interpretation and, importantly, enable fusion/coregistration with different SAR images or other types of geospatial data. However, objects not included in the terrain model used for projection -- i.e. buildings and vegetation -- still appear distorted. In the third and fourth columns of \cref{fig:geometries}, the two SAR images alongside their respective image-specific heights in ground range projection are shown. The Eiffel Towers are now pointing in different directions (towards the respective sensor), but ground pixels appear at the same position in both of the images, enabling pixel-by-pixel superimposition. We refer to the heights projected from slant range geometry onto the DTM as heights in image geometry or still as \textit{slant} heights since these height values originate from slant range and retain the characteristics of the original geometry being image-specific. The last column in \cref{fig:geometries} displays the corresponding heights in a map projection, which is independent of the image geometry and thus identical for both SAR images shown.

\vspace{1mm}
\noindent\textbf{Dataset Limitations and Challenges}
The dataset utilized in this study presents several limitations that may introduce uncertainties into the model's performance. First, there are inconsistencies in the acquisition dates of different images capturing the same geographic scene, meaning that physical or structural changes could have occurred in the intervening period. For instance, a newly built building visible in only one of the provided views could confuse the model. Additionally, discrepancies between the acquisition dates of images and corresponding ground truth data -- whether LiDAR or building footprint annotations -- introduce errors both in training and validation, as both imagery and ground truth may not accurately reflect the same spatial conditions.
Furthermore, the digital terrain model used for terrain correction bears inherent inaccuracies, with deviations that can reach several meters, particularly in densely constructed urban areas. These inaccuracies propagate into geolocation errors, potentially distorting spatial alignment between SAR images captured from varying perspectives and with respect to the ground truth data. 

\vspace{1mm}
\noindent\textbf{Parameter Preprocessing}
The acquisition parameters, which are incorporated into the transformer via the APE module, undergo preprocessing to enhance their interpretability for the neural network, as described in \cref{eq:preprocessing}. To preserve its cyclic nature and avoid artificial discontinuities, the azimuth angle $Az$, which spans from $0^\circ$ to $360^\circ$, is represented using its sine and cosine components. The looking angle $\theta$ is transformed using the cotangent function, as this effectively approximates the ratio between building height and the corresponding layover extent on a tangential plane. The imaging mode $m$ is mapped to a single-digit identifier. While the imaging mode could alternatively be substituted by the sensor's resolution -- given their direct correlation -- it is included here as an example of how semantic or non-numerical metadata, such as sensor type, input modality, or polarization, can be incorporated into the model.

\section{Additional Information: Metrics}

\noindent\textbf{Segmentation Metrics}
To evaluate the performance of the binary building footprint segmentation, we used the overall accuracy (OA)
\begin{equation}
    \text{OA} = \frac{TP + TN}{TP + TN + FP + FN} 
\end{equation}
with $TP$ and $FP$ the true and false positives, and $TN$ and $FN$ the true and false negatives, and the mean Intersection over Union (mIoU):
\begin{equation}
    \text{mIoU} = \frac{1}{2} \biggl( \frac{TP}{TP + FP + FN} + \frac{TN}{TN + FP + FN} \biggr).
\end{equation}

\vspace{1mm}
\noindent\textbf{Regression Metrics}
To assess the performance of the height reconstruction (both in map and image geometries), we utilized the Mean Absolute Error (MAE)
\begin{equation}
    \text{MAE} = \frac{1}{n} \sum_{i=1}^{n} |y_i - \hat{y}_i|,
\end{equation}
with $n$ as the number of data points, $y_i$ is the actual target value for pixel $i$, and $\hat{y}_i$ as the predicted value for pixel $i$, the Root Mean Squared Error (RMSE)
\begin{equation}
    \text{RMSE} = \sqrt{\frac{1}{n} \sum_{i=1}^{n} (y_i - \hat{y}_i)^2},
\end{equation}
and the Structural Similarity Index Measure (SSIM) \cite{Wang2004ssim} as some sort of relative metric, which is intended to reflect people's perception:
\begin{equation}
	\text{SSIM} = \frac{(2\mu_y \mu_{\hat{y}} + C_1)(2\sigma_{y\hat{y}} + C_2)}{(\mu_y^2 + \mu_{\hat{y}}^2 + C_1)(\sigma_y^2 + \sigma_{y\hat{y}}^2 + C_2)},
\end{equation}
where $y$ and $\hat{y}$ are the predicted and target images, $\mu_y$ and $\mu_{\hat{y}}$ their average pixel intensities,  $\sigma_y^2$ and $\sigma_{y\hat{y}}^2$ the corresponding variances, $\sigma_{y\hat{y}}$ representing the covariance, and $C_1$ and $C_2$ as constants to numerically stabilize the division.

\begin{table*}[t!]
    \centering
    \caption{Numerical results from fully-supervised experiments (no pre-training) using the \textbf{ViT-Base backbone} in the 2-view scenario. Performance gains achieved with the APE module are comparable to those reported for the ViT-Large configuration (refer to the main paper).}
    \begin{tabular}{cccccccccc}
    \toprule
    \multicolumn{2}{c}{} & \multicolumn{2}{c}{\textbf{\makecell{Classification \\ Footprints}}} & \multicolumn{3}{c}{\textbf{\makecell{Regression \\ Height}}} & \multicolumn{3}{c}{\textbf{\makecell{Regression \\ Height (Slant)}}}\\
    \cmidrule(lr){3-4} \cmidrule(lr){5-7} \cmidrule(lr){8-10} 
    \# Views & Model & mIoU & OA & MAE & RMSE & SSIM & MAE & RMSE & SSIM \\
    \cmidrule(lr){1-10} 
    \multirow{2}{*}{2} & DPT-Base \cite{Ranftl2021dpt} & \bm{$0.73$} & \bm{$0.92$} & $4.86$ & $7.23$ & \bm{$0.85$} & $5.06$ & $7.38$ & \bm{$0.88$} \\
                       & SARFormer (ours) & \bm{$0.73$} & \bm{$0.92$} & \bm{$4.44$} & \bm{$6.68$} & \bm{$0.85$} & \bm{$4.69$} & \bm{$6.94$} & \bm{$0.88$} \\
    \arrayrulecolor{black}\bottomrule
    \end{tabular}
    \label{tab:fromScratch_base}
\end{table*}

\section{Additional Information: Training}

\noindent\textbf{Loss Function}
Since we are dealing with a long-tailed distribution, characterized by a substantial number of ground pixels at a height of zero, the models usually tend to underestimate heights. This observation underscores the rationale for employing an asymmetric loss function:
\begin{equation}
    l_{\text{asym}} = \frac{1}{n} \sum_{i=1}^{n} 
    \begin{cases}
        w_{\text{underestimated}} \cdot |\hat{y}_i - y_i|, & \text{if } \hat{y}_i < y_i \\
        w_{\text{overestimated}} \cdot |\hat{y}_i - y_i|, & \text{if } \hat{y}_i \geq y_i
    \end{cases}
\end{equation}
with $\hat{y}$ the predictions and $y$ the target, $n$ the number of pixels, $w_{\text{underestimated}} = 1.5$ and $w_{\text{overestimated}} = 1$.
To penalize errors along edges and to further improve small details, the gradient loss 
\begin{equation}
    l_{\text{grad}} = \sum_{d \in \{x, y\}} \left\| \nabla_d(\hat{y}) - \nabla_d(y) \right\|_1
\end{equation}
where $\nabla_d$ is the spatial derivative in the dimension $d$ (determined through Sobel operator), and normal loss is added (details can be found in \cite{hu2018loss}):
\begin{equation}
    l_{\text{normal}} = \frac{1}{n} \sum_{i=1}^{n} \left( 1 - \frac{\langle \vec{n}_i^{\hat{y}}, \vec{n}_i^{y} \rangle}{\sqrt{\langle \vec{n}_i^{\hat{y}}, \vec{n}_i^{\hat{y}} \rangle} \sqrt{\langle \vec{n}_i^{y}, \vec{n}_i^{y} \vphantom{\vec{n}_i^{\hat{y}}} \rangle}} \right)
\end{equation}
with $\scriptstyle{\vec{n}_i^{\text{x}} = \left[ \begin{smallmatrix} -\nabla_x(\text{x}_i), & -\nabla_y(\text{x}_i), & 1 \end{smallmatrix} \right]^{\top}, \; \text{x} \in \{y, \hat{y}\}}$ and $\langle \cdot, \cdot \rangle$ the inner product of vectors. The combined loss function for the regression task is the weighted sum of the losses above:
\begin{equation}
    \mathcal{L}_{\text{regression}} = \alpha \cdot l_{\text{asym}} + \beta \cdot l_{\text{norm}} + \gamma \cdot l_{\text{grad}}
\end{equation}
with $\alpha = \beta = 1$ and $\gamma = 0.1$.
For the segmentation task, i.e. the building footprints, the binary cross entropy loss is used, which can be expressed as (using logits $\hat{y}$):
\begin{multline}
    \mathcal{L}_{\text{segmentation}} = l_{\text{BCE}} = \\
    - \frac{1}{n} \sum_{i=1}^{n} \left( y_i \log(\sigma(\hat{y}_i)) + (1 - y_i) \log(1 - \sigma(\hat{y}_i)) \right)
\end{multline}
where \( \sigma(x) = \frac{1}{1 + e^{-x}} \) is the sigmoid function. The total loss function is formulated as a weighted sum of the individual task-specific losses:
\begin{equation}
    \mathcal{L}_{\text{MTL}} = \mathcal{L}_{\text{height}} + \mathcal{L}_{\text{height slant}} + 0.1 \cdot \mathcal{L}_{\text{buildings}}.
\end{equation}

\vspace{1mm}
\noindent\textbf{Multitask DPT Setup}
We adopt the dense prediction strategy using ViT-based models as proposed by \cite{Ranftl2021dpt} (DPT). The projection of features derived from multiple views and their associated metadata tokens is described in detail in the main manuscript. For the 12-layer \textit{ViT-Base} configuration, features are extracted after layers 3, 6, 9, and 12. For the \textit{ViT-Large} configuration, features are extracted after layers 5, 12, 18, and 24. Similarly, for the \textit{ViT-Huge} configuration, features are extracted after layers 8, 16, 24, and 32.

To infer annotations for the three tasks in the multitask framework, we extend the final convolutional block of the DPT with three task-specific blocks. Each block consists of five convolutional layers, each followed by a \textit{LeakyReLU} activation function.

\begin{figure}
    \centering
    \includegraphics[width=\linewidth]{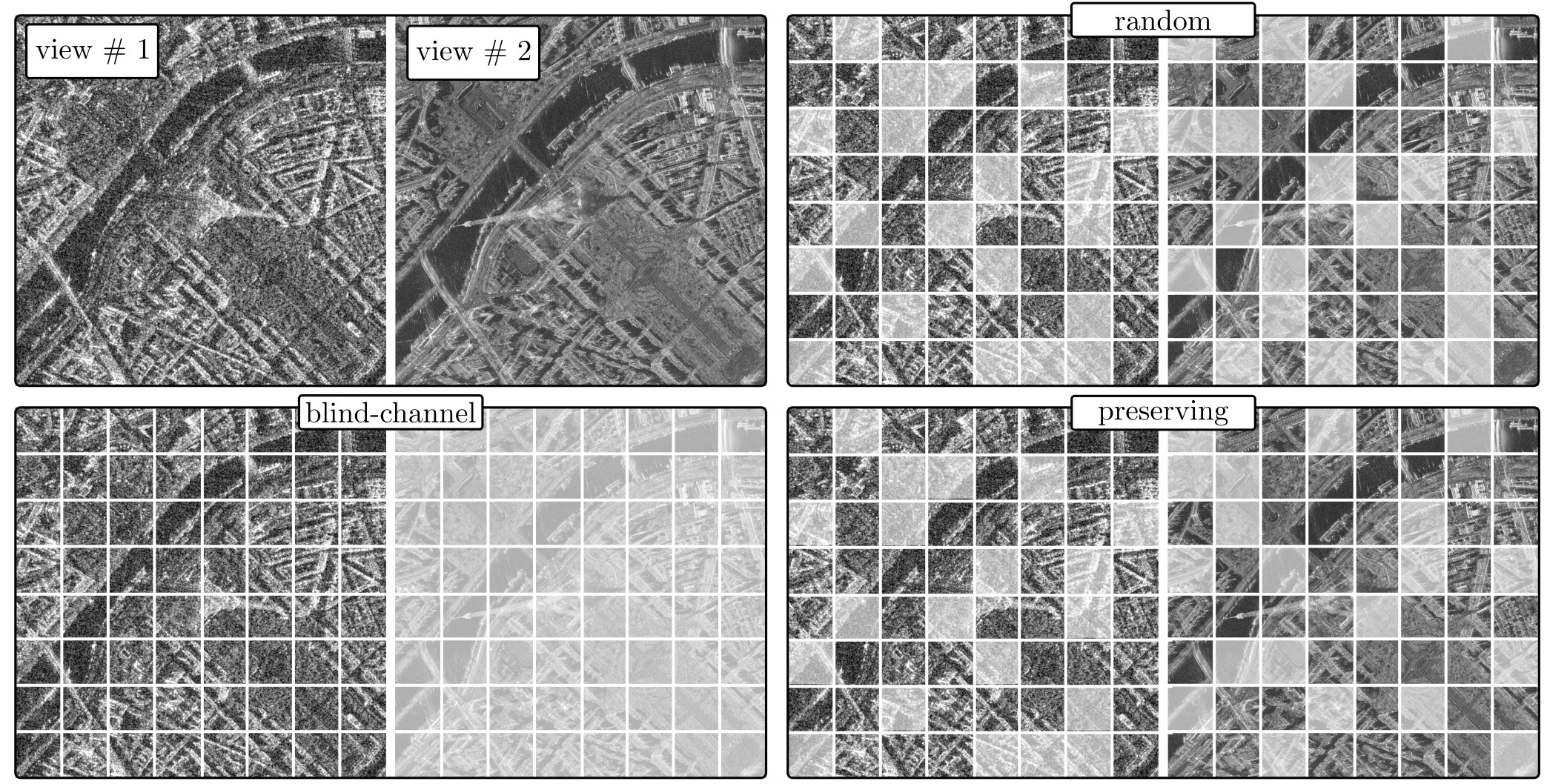}
    \caption{Different masking strategies for multi-view scenarios. The \textit{preserving} masking strategy (lower right) ensures that at least one view remains active for all locations across views \#1 and \#2. In contrast, the \textit{random} masking strategy (upper right) does not guarantee this consistency. The \textit{blind-channel} masking strategy (lower left) is a special case of \textit{preserving}, where one view is entirely masked while the other remains fully active.}
    \label{fig:masking}
\end{figure}

\vspace{1mm}
\noindent\textbf{Masking strategies for MAE pre-training}
In the context of remote sensing images, traditional masking strategies employed in masked autoencoders must be rethought to address the fundamental differences from natural object-centered images. Unlike photographs of objects (such as dogs for instance), where masking significant portions still allows recognition due to the structured and object-centric nature of the image, remote sensing images often lack such intrinsic coherence. 
For example, if a single building within a forested area is masked, it is basically impossible for the model to reconstruct it due to the absence of sufficient contextual information.
To address this issue, we introduce novel masking strategies tailored to remote sensing data. Our approach leverages multiple views of the same scene captured under varying acquisition conditions, such as differing resolutions, angles, and directions. By ensuring that masking does not occlude the same patch across all views, the model retains at least one perspective for reference, thereby enhancing reconstruction potential while keeping the intrinsic complexity arising from acquisition variability. We refer to this strategy as \textit{preserving}. A notable extreme of this strategy, termed \textit{blind-channel} masking, involves completely masking one view, challenging the MAE to reconstruct it solely from the unmasked complementary view.
The \textit{blind-channel} scenario necessitates encoding the acquisition parameters (as being done by the APE module) since these cannot be inferred from the data in the masked view.
These strategies exploit the rich heterogeneity of remote sensing data, fostering more robust and semantically meaningful representations. Compare \cref{fig:masking} for a visual example of the different strategies.

\begin{table}[t]
    \centering
    \caption{Comparison between different model size configurations (on a subset of the metrics). The setting was chosen to the best-performing: two views, active APE module, pre-trained using the \textit{preserving} strategy.}
    \begin{tabular}{lccc}
    \toprule
    \textbf{Model Size} & \textbf{mIoU} & \textbf{MAE} (map) &  \textbf{MAE} (slant) \\
    \cmidrule(lr){1-1} \cmidrule(lr){2-2} \cmidrule(lr){3-3} \cmidrule(lr){4-4}
    ViT-Base & $0.73$ & $4.26$ & $4.39$ \\
    ViT-Large & $0.74$ & $4.12$ & \bm{$3.96$} \\
    ViT-Huge & \bm{$0.76$} & \bm{$4.04$} & \bm{$3.96$} \\
    \arrayrulecolor{black}\bottomrule
    \end{tabular}
    \label{tab:model_size}
\end{table}

\section{Additional Results}

\noindent\textbf{Effect of Backbones}
As discussed in the main paper, we trained our best-performing configuration ($2$ views, active APE, and \textit{preserving} masking during pre-training) using three different backbone architectures: ViT-Base, ViT-Large, and ViT-Huge. \cref{tab:model_size} presents a subset of evaluation metrics on the test set for these configurations, demonstrating a consistent trend where larger model sizes lead to improved performance.
Furthermore, \cref{tab:fromScratch_base} illustrates the effect of incorporating the APE module into the ViT-Base backbone, evaluated in the 2-view setup. Notably, the inclusion of the \textit{metatoken} demonstrates significant benefits, particularly for the height estimation task.

\begin{figure}
    \centering
    \includegraphics[width=\linewidth]{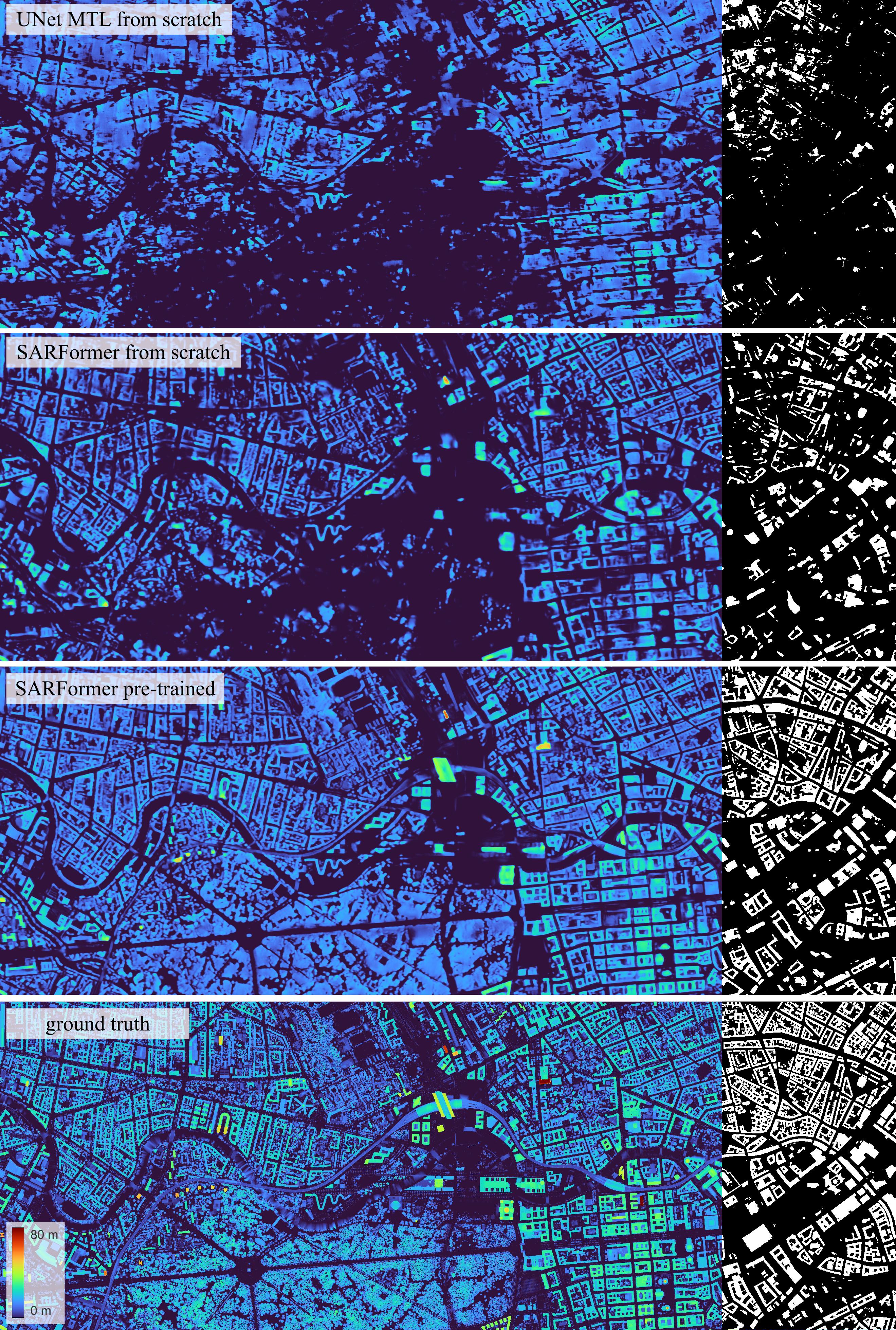}
    \caption{Next to the ground truth (bottom), we present model outputs trained on an extremely limited dataset consisting of only two SM images of Paris. Inference was conducted on two HS images of Berlin captured from different viewing angles than those used in training. Notably, the pre-trained \textit{SARFormer} (third row) demonstrates the highest resilience to this multifactorial domain shift, encompassing location, resolution, and geometric differences. For comparison, we also display results from UNet and \textit{SARFormer} trained from scratch (first and second rows, respectively).}
    \label{fig:pretrained_fewlabels}
\end{figure}

\vspace{1mm}
\noindent\textbf{Fine-Tuning on limited Labels}
To further highlight the effectiveness of the proposed pre-training paradigm, we minimized the labeled dataset for fine-tuning to just two SM images from a single location, Paris. This setup introduces a significant domain shift in multiple regards during testing, as it includes data from different locations, acquisition modes, and looking angles. \cref{fig:pretrained_fewlabels} presents a visual comparison of outputs -- specifically, height maps and building footprints -- generated by a UNet (trained from scratch), the non-pre-trained \textit{SARFormer}, and the pre-trained \textit{SARFormer}. Two HS scenes from Berlin served as model input. Although performance remains below that achieved with the full dataset, the benefits of pre-training are evident, underscoring its value, particularly for few-label or out-of-domain scenarios. \cref{tab:limited_labels} shows the error metrics for the entire test set (the same as all other experiments were evaluated on) in the limited-label scenario.

\begin{figure*}[b]
    \centering
    \includegraphics[width=\linewidth]{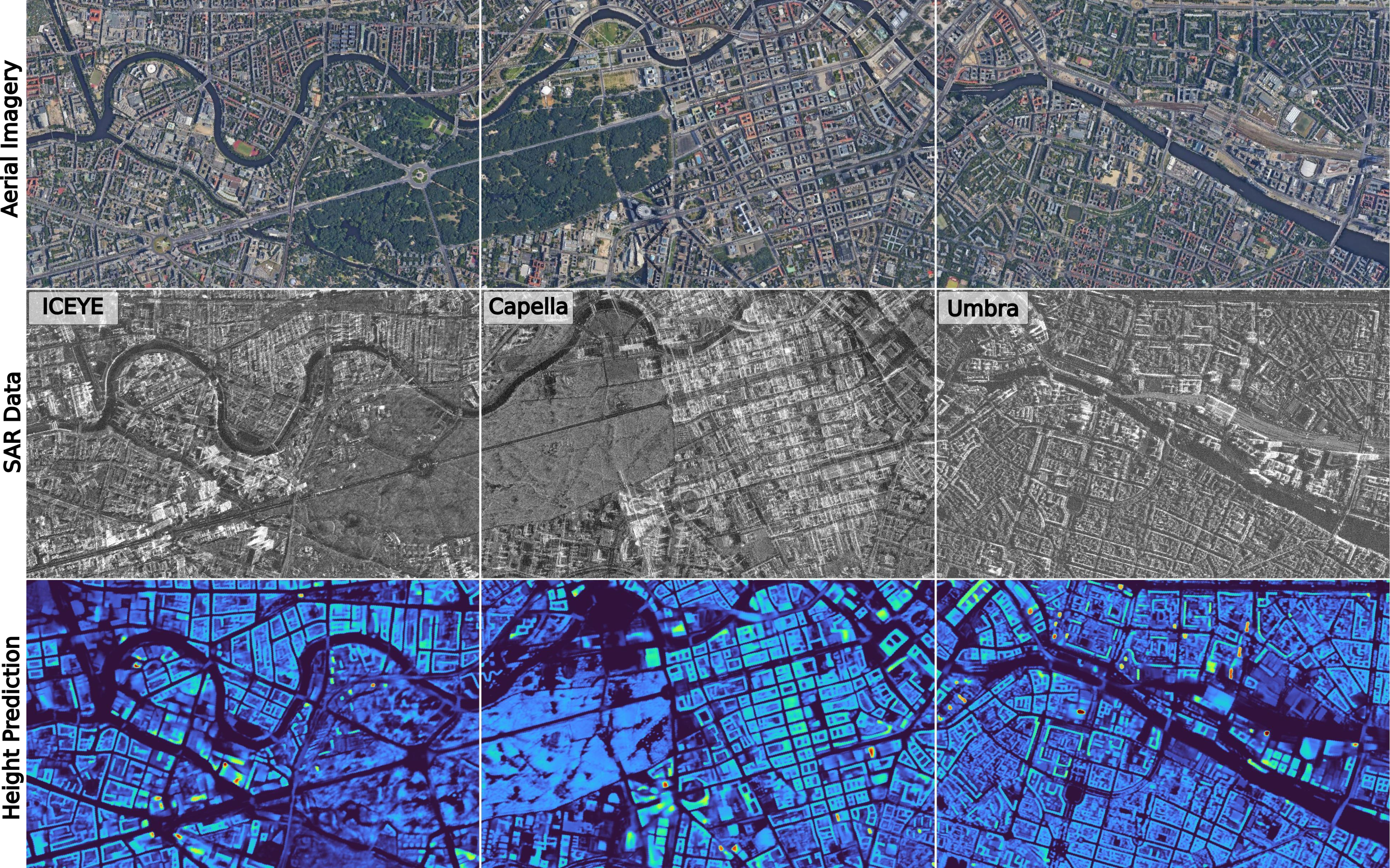}
    \caption{Three SAR spotlight acquisitions over the city of Berlin, provided by \textit{ICEYE}, \textit{Capella Space}, and \textit{Umbra}, were used to infer the \textit{SARFormer} model. Despite differences in acquisition characteristics, the combined nDSMs (bottom row) result in a homogeneous and coherent reconstruction. Aerial imagery is taken from Google, \copyright 2025.}
    \label{fig:newspace}
\end{figure*}

\section{Additional Visual Results}

\begin{figure*}
    \centering
    \includegraphics[width=\linewidth]{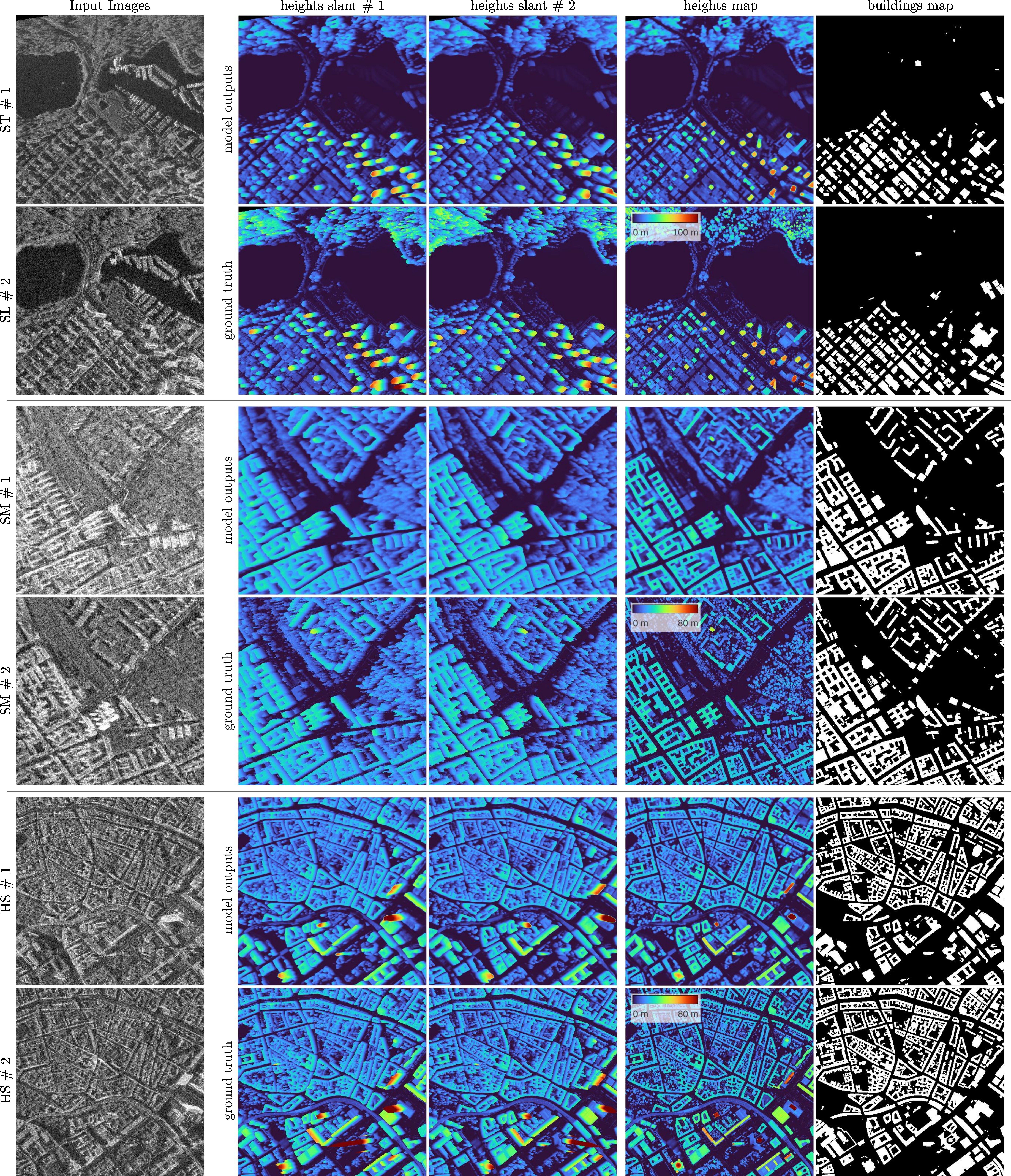}
    \caption{Model outputs for three scenes generated by the pre-trained \textit{SARFormer} (ViT-Huge) using $2$ input views. The upper rows present the model's predictions for the three downstream tasks, while the lower rows display the corresponding ground truth data.}
    \label{fig:example_outputs}
\end{figure*}

\noindent\textbf{Demonstration of best-performing configuration}
\cref{fig:example_outputs} presents the outputs of all three tasks on three representative SAR scenes. These results were generated using the pre-trained \textit{SARFormer} with the ViT-Huge backbone, activated APE module, and the \textit{preserving} masking strategy during pre-training. For each example, the top row shows the model outputs, while the bottom row displays the corresponding ground truth data. The depicted scenes are from Vancouver and Berlin, both of which were entirely excluded from the training set.
In the first example, two different imaging modes were utilized to reconstruct a complex scene containing multiple high-rise buildings. The results, in both slant and map geometries, are closely aligned with the ground truth. 
The second example illustrates a challenging case involving SM input data, characterized by low spatial resolution and high noise levels. While the performance is inferior compared to Spotlight images, the model still achieves acceptable results.
An intriguing detail in the final example is the absence of the Berlin TV Tower in the model’s prediction, which is distinctly visible as the tallest structure in the ground truth. This omission is very likely due to the weak radar response of the tower. Only the sphere at the top of the structure is faintly discernible in the SAR images, a feature detectable only by trained observers. Here, the methodology reaches its physical limitations.

\vspace{1mm}
\noindent\textbf{Extension to other missions}
Although a detailed description of integrating various satellite missions into the \textit{SARFormer} framework is beyond the scope of this manuscript, it is important to note that such integration is straightforward. \Cref{fig:newspace} presents an exemplary output from a \textit{SARFormer} variant that was pre-trained and fine-tuned on an extended version of the dataset described here. This extended dataset includes imagery from \textit{ICEYE}, \textit{Umbra}, and \textit{Capella Space} in addition to the previously mentioned \textit{TerraSAR-X} data. The only modification relative to the manuscript was to replace the encoding of a discretized imaging mode $m$ with the encoding of the azimuth and range resolutions of the corresponding product since the nomenclature of the imaging modes differs between providers.

\begin{figure*}
    \centering
    \includegraphics[width=.91\linewidth]{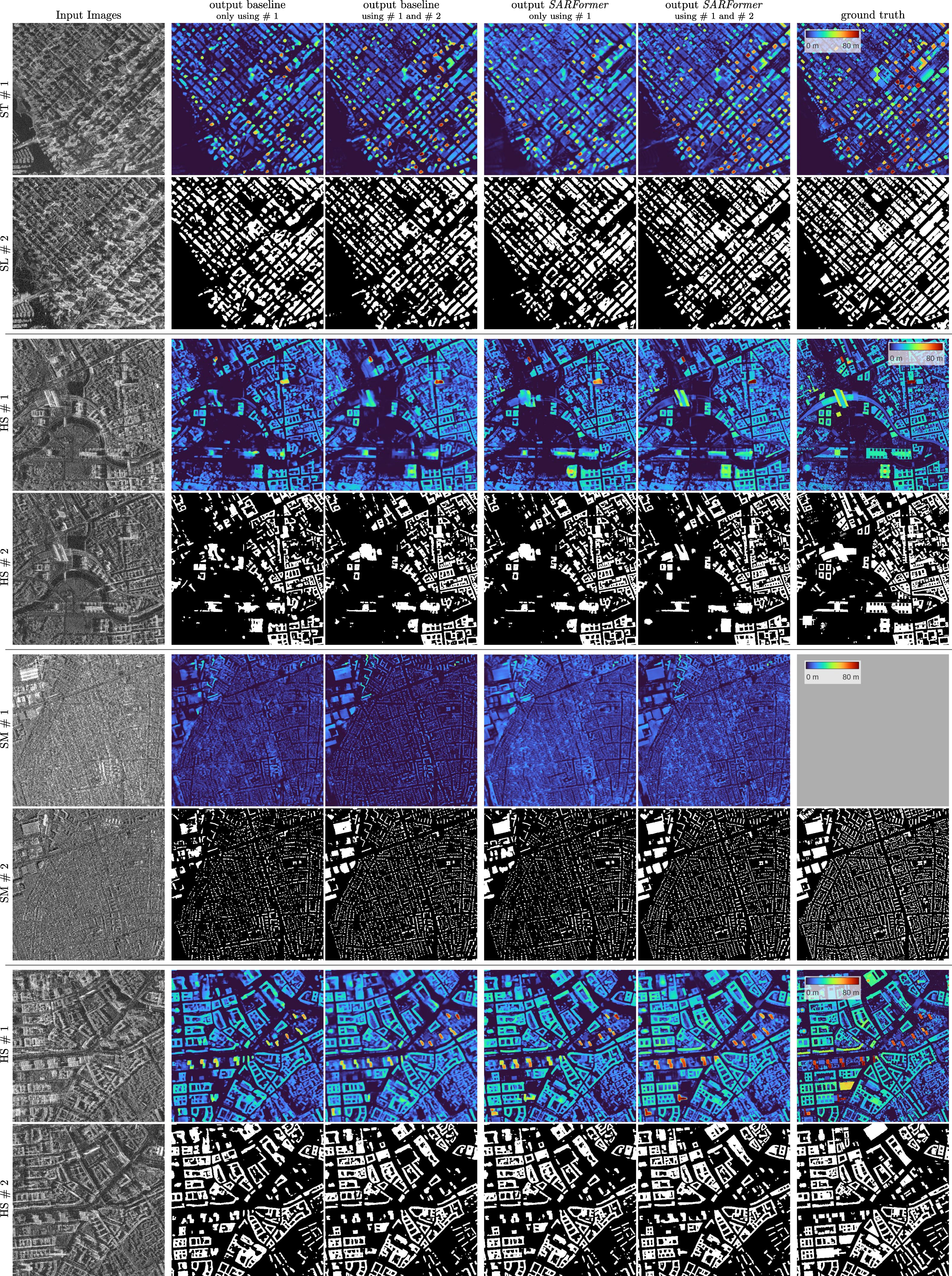}
    \caption{Comparison of SARFormer and baseline models in both single-view and two-view scenarios. The final column displays the ground truth data. Note that height labels were unavailable for the third scene.}
    \label{fig:comparison_baseline}
\end{figure*}

\vspace{1mm}
\noindent\textbf{Comparison to Baseline}
\cref{fig:comparison_baseline} compares the outputs of four different models with the corresponding ground truth. The baseline is a UNet in the multi-task configuration, evaluated for both single-view and two-view scenarios. In contrast, we include results from the pre-trained \textit{SARFormer} (ViT-Large), also evaluated for single-view and two-view cases.  
The performance comparison across the four illustrated scenes highlights several key observations. First, the addition of a second view significantly enhances the reconstruction capability of the models, both in terms of height accuracy and building shapes. Furthermore, SARFormer demonstrates superior performance compared to the baseline, in both single-view and two-view scenarios. 
For the third scene, no DSM as ground truth was available, so the comparison is limited to building footprints. In this case, the \textit{SARFormer} models again produced results that align more closely with the labels compared to the baseline models.  
Overall, it was observed that the proposed \textit{SARFormer} architecture particularly excels in complex scenarios, such as those involving low-resolution data, small structures, or heavily mixed layover signals.

\end{document}